%% file: main.tex
\pdfoutput=1
\documentclass{article}
\usepackage[preprint]{neurips_2026}

\usepackage[utf8]{inputenc} %
\usepackage[T1]{fontenc}    %
\usepackage{hyperref}       %
\usepackage{url}            %
\usepackage{booktabs}       %
\usepackage{amsfonts}       %
\usepackage{nicefrac}       %
\usepackage{microtype}      %
\usepackage{xcolor}         %
\usepackage{graphicx}
\usepackage{amsmath}
\usepackage{multirow}
\usepackage{algorithm}
\usepackage{algorithmic}
\usepackage{amssymb}
\usepackage{mathtools}
\usepackage{amsthm}
\usepackage{bm}

\usepackage{wrapfig}
\usepackage{caption}
\usepackage{tikz}
\hypersetup{hidelinks}
\usetikzlibrary{arrows.meta,calc,decorations.pathreplacing}
\title{Disentangled Representation Learning via Flow Matching}

\author{
  Jinjin Chi$^{1,2,3}$ \qquad Taoping Liu$^{1}$ \qquad Mengtao Yin$^{1}$\qquad Ximing Li$^{1,3}\thanks{Corresponding author.}$ \qquad Yongcheng Jing$^{2}$  \\ \textbf{Leszek Rutkowski}$^{4,5,6}$ \qquad \textbf{Jialie Shen}$^{7}$\qquad \textbf{Dacheng Tao}$^{2}$\\
  $^1$College of Computer Science and Technology, Jilin University \\
  $^2$College of Computing and Data Science, Nanyang Technological University\\
  $^3$Key Laboratory of Symbolic Computation and Knowledge Engineering, MoE, Jilin University\\
  $^{4}$Systems Research Institute, Polish Academy of Sciences\\
  $^{5}$AGH University of Krakow\\
  $^{6}$SAN University\\
  $^7$City St George's, University of London\\
  \{\texttt{chijinjin616@gmail.com}, \texttt{liximing86@gmail.com}\}
  }

\begin{document}

\maketitle

\begin{abstract}
  Disentangled representation learning aims to capture the underlying explanatory factors of observed data, enabling a principled understanding of the data-generating process. Recent advances in generative modeling have introduced new paradigms for learning such representations. However, existing diffusion-based methods encourage factor independence via inductive biases, yet frequently lack strong semantic alignment. In this work, we propose a flow-matching–based framework for disentangled representation learning, which casts disentanglement as learning factor-conditioned flows in a compact latent space. To encourage semantic separation among factor representations, we introduce a non-overlap (orthogonality) regularizer that suppresses cross-factor interference and reduces information leakage between factors. Extensive experiments across multiple datasets demonstrate consistent improvements over representative baselines, yielding higher disentanglement scores as well as improved controllability and sample fidelity.
\end{abstract}

\input{introduction}

\input{related-work}
\input{background}

\input{method}

\input{experiments}

\section{Conclusion}
We present a flow-matching framework for disentangled representation learning that models generation as factor-conditioned transport in a compact latent space. An orthogonality regularizer discourages overlap
among token-associated velocity components, encouraging semantic separation among the learned factor representations. Experiments show improved mean DCI-D scores on the synthetic benchmarks and lower FID on CelebA compared with the evaluated diffusion-based baselines. Factor-swapping results further illustrate semantic controllability. These findings support structured latent transport as a promising approach to learning disentangled and controllable representations.

\begin{ack}
This research is supported by the Scientific and Technological Developing Scheme of Jilin Province (No. 20250102213JC), the National Natural Science Foundation of China (NSFC) (Nos. 62006094 and 62276113), and the National Science Centre, Poland (NCN) (No. 2025/59/B/ST6/01950). It is also supported by the RIE2025 Industry Alignment Fund – Industry Collaboration Projects (IAF-ICP) (Award I2301E0026), administered by A*STAR, as well as by Alibaba Group and NTU Singapore through Alibaba-NTU Global e-Sustainability CorpLab (ANGEL).
\end{ack}

\bibliographystyle{plainnat}
\bibliography{ref}

\appendix
\section{Broader Societal Impact}
While this work aims to advance disentangled representation learning and controllable generative modeling, enabling applications in image editing, data analysis, and scientific understanding, it also raises potential concerns regarding misuse, such as unauthorized content manipulation or the creation of misleading media. Addressing potential biases in learned representations and ensuring fair and responsible deployment remain important directions for future research.

\section{Limitations}
Our method uses a prespecified number of factor tokens $N$ and does not automatically determine the required capacity. When $N$ is below the ground-truth factor count, a dedicated token for each factor is impossible, which may limit token-level separation. When $N$ is over-specified, output attention provides soft routing without explicitly pruning unnecessary components. Our experiments show that additional capacity can preserve or improve factor predictability while reducing completeness on Shapes3D and MPI3D-toy. Selecting $N$ therefore remains an empirical design choice, motivating future work on adaptive token allocation and pruning.

\section{More Implementation Details}

\subsection{Algorithm}
Algorithm~\ref{althm1} summarizes the training procedure for the proposed framework.

\begin{algorithm}[h]
\caption{Overall Disentangled Representation Learning Framework}
\label{althm1}
\begin{algorithmic}[1]
\STATE \textbf{Input}: pretrained VQ-GAN encoder $E(\cdot)$, networks $T_{\gamma}$ and $\mathbf v_{\theta}$, number of tokens $N$, orthogonality weight $\lambda_{\mathrm{orth}}$
\STATE Randomly initialize $\theta$ and $\gamma$
\WHILE{not converged}
    \STATE Sample an image $I$ from the training set
    \STATE Encode the image: $\mathbf z_1 \leftarrow E(I)$
    \STATE Extract factor tokens: $S_\gamma(I)=\{\mathbf s_i\}_{i=1}^{N} \leftarrow T_\gamma(I)$
    \STATE Sample $\mathbf z_0 \sim \mathcal N(0,\mathbf I)$ and $t \sim \mathcal U[0,1]$
    \STATE Construct $\mathbf z_t \leftarrow (1-t)\mathbf z_0 + t\mathbf z_1$
    \STATE Compute target velocity: $\mathbf u \leftarrow \mathbf z_1-\mathbf z_0$
    \STATE Predict aggregate velocity using cross-attention conditioning:
    $\widehat{\mathbf v} \leftarrow \mathbf v_\theta(\mathbf z_t,S_\gamma(I),t)$
    \STATE Compute output-attention weights $\{\mathrm{Attn}_i\}_{i=1}^{N}$ using Eq.~\eqref{eq:token-attn}
    \STATE Obtain factor-specific velocities:
    $\mathbf v_\theta^{(i)} \leftarrow \mathrm{Attn}_i \odot \widehat{\mathbf v}$
    \STATE Compute $\mathcal L_{\mathrm{FM}} \leftarrow \|\widehat{\mathbf v}-\mathbf u\|_2^2$
    \STATE Compute $\mathcal L_{\mathrm{orth}}$ using Eq.~\eqref{eq:orth-regularizer}
    \STATE Set $\mathcal L \leftarrow \mathcal L_{\mathrm{FM}}+\lambda_{\mathrm{orth}}\mathcal L_{\mathrm{orth}}$
    \STATE Update $(\theta,\gamma)$ using $\nabla_{\theta,\gamma}\mathcal L$
\ENDWHILE
\STATE \textbf{Output}: $\theta,\gamma$
\end{algorithmic}
\end{algorithm}

\subsection{Model Architecture}
We adopt the factor encoder architecture used in the official implementation of EncDiff~\citep{yang2024diffusion}. The encoder consists of a convolutional backbone with progressive downsampling and residual blocks, followed by a fully connected layer that produces $N$ scalar representations. Each scalar is mapped to a 16-dimensional factor token by a separate MLP, with no parameter sharing across factors.
Table~\ref{tab:encoder} summarizes the architecture.

\begin{table}[h]
\centering
\caption{Architecture of the factor encoder $T_\gamma$ (input $64{\times}64{\times}3$).}
\label{tab:encoder}
\small
\begin{tabular}{ll}
\toprule
Layer & Configuration \\
\midrule
Conv $4{\times}4$, $3\to128$, stride 2 & BN, ReLU \\
Conv $4{\times}4$, $128\to128$, stride 2 & BN, ReLU \\
Conv $4{\times}4$, $128\to128$, stride 2 & BN, ReLU \\
Conv $4{\times}4$, $128\to128$, stride 2 & BN, ReLU \\
2$\times$ residual block, 128 ch & Conv $3{\times}3$ + Conv $1{\times}1$, BN, ReLU \\
Flatten & $128\times4\times4 = 2048$ \\
FC $2048\to N$ & factor scalars $c_1,\dots,c_N$ \\
\midrule
Per-factor MLP (not shared across $i$) & $c_i$: FC $1\to64$, ELU, FC $64\to128$, ELU, FC $128\to16$ \\
Output & $S_\gamma(I)=\{\mathbf s_i\}_{i=1}^{N}$, $\mathbf s_i\in\mathbb R^{16}$ \\
\bottomrule
\end{tabular}
\end{table}

For velocity prediction, we use a U-Net with a shared architecture across Shapes3D, Cars3D, and MPI3D-toy. Table~\ref{tab:unet-config} summarizes the architecture and dataset-specific training settings.

\begin{table}[h]
\centering
\caption{U-Net architecture and training settings for velocity prediction on the three synthetic datasets.}
\label{tab:unet-config}
\small
\begin{tabular}{ll}
\toprule
Parameter & Shapes3D / Cars3D / MPI3D-toy \\
\midrule
Input latent (VQ-GAN, $f{=}4$, 2048 codes) & $16\times16\times3$ \\
Model channels & 64 \\
Channel multipliers & [1, 2, 4, 4] \\
Res.\ blocks per level & 2 \\
Attention resolutions & $16, 8, 4$ \\
Attention heads & 8 \\
Cross-attention context dim & 16 ($=\dim \mathbf s_i$) \\
Number of tokens $N$ & 20 \\
Optimizer & AdamW \\
Batch size & 128 \\
Dataset size & 0.48M / 0.018M / 1.04M \\
\bottomrule
\end{tabular}
\end{table}

\subsection{More Experiments}
\label{app:additional_results}

\paragraph{Informativeness and completeness under over-specification.}
To complement the sensitivity analysis, we report DCI informativeness (DCI-I) and completeness (DCI-C) for $N\in\{10,20,30\}$. These settings exceed the ground-truth factor counts of all three datasets. DCI-I measures factor predictability, while DCI-C measures the concentration of each factor's predictive importance across representation components.

\begin{table}[!t]
\centering
\footnotesize
\setlength{\tabcolsep}{4pt}
\caption{DCI-I and DCI-C under over-specification of $N$ (mean$\pm$std).}
\label{tab:dci_ic_sensitivity}
\begin{tabular}{lcccccc}
\toprule
\textbf{Dataset}
& \multicolumn{2}{c}{$N=10$}
& \multicolumn{2}{c}{$N=20$}
& \multicolumn{2}{c}{$N=30$} \\
\cmidrule(lr){2-3}
\cmidrule(lr){4-5}
\cmidrule(lr){6-7}
& DCI-I $\uparrow$ & DCI-C $\uparrow$
& DCI-I $\uparrow$ & DCI-C $\uparrow$
& DCI-I $\uparrow$ & DCI-C $\uparrow$ \\
\midrule
Cars3D
& $0.698 \pm 0.136$ & $0.251 \pm 0.027$
& $0.813 \pm 0.101$ & $0.279 \pm 0.012$
& $0.814 \pm 0.110$ & $0.289 \pm 0.014$ \\
Shapes3D
& $0.999 \pm 0.000$ & $0.791 \pm 0.017$
& $0.999 \pm 0.000$ & $0.699 \pm 0.038$
& $0.999 \pm 0.000$ & $0.693 \pm 0.043$ \\
MPI3D-toy
& $0.815 \pm 0.010$ & $0.609 \pm 0.023$
& $0.829 \pm 0.036$ & $0.584 \pm 0.019$
& $0.859 \pm 0.004$ & $0.573 \pm 0.064$ \\
\bottomrule
\end{tabular}
\end{table}

Table~\ref{tab:dci_ic_sensitivity} shows that DCI-I remains unchanged on Shapes3D and increases on MPI3D-toy and Cars3D as $N$ increases. DCI-C decreases on Shapes3D and MPI3D-toy but increases on Cars3D. The results on Shapes3D and MPI3D-toy are consistent with existing factor information becoming less concentrated across components while remaining predictable. They do not establish whether individual tokens are inactive or redundantly used.

\begin{figure*}[!t]
\centering
\includegraphics[width=0.9\textwidth]{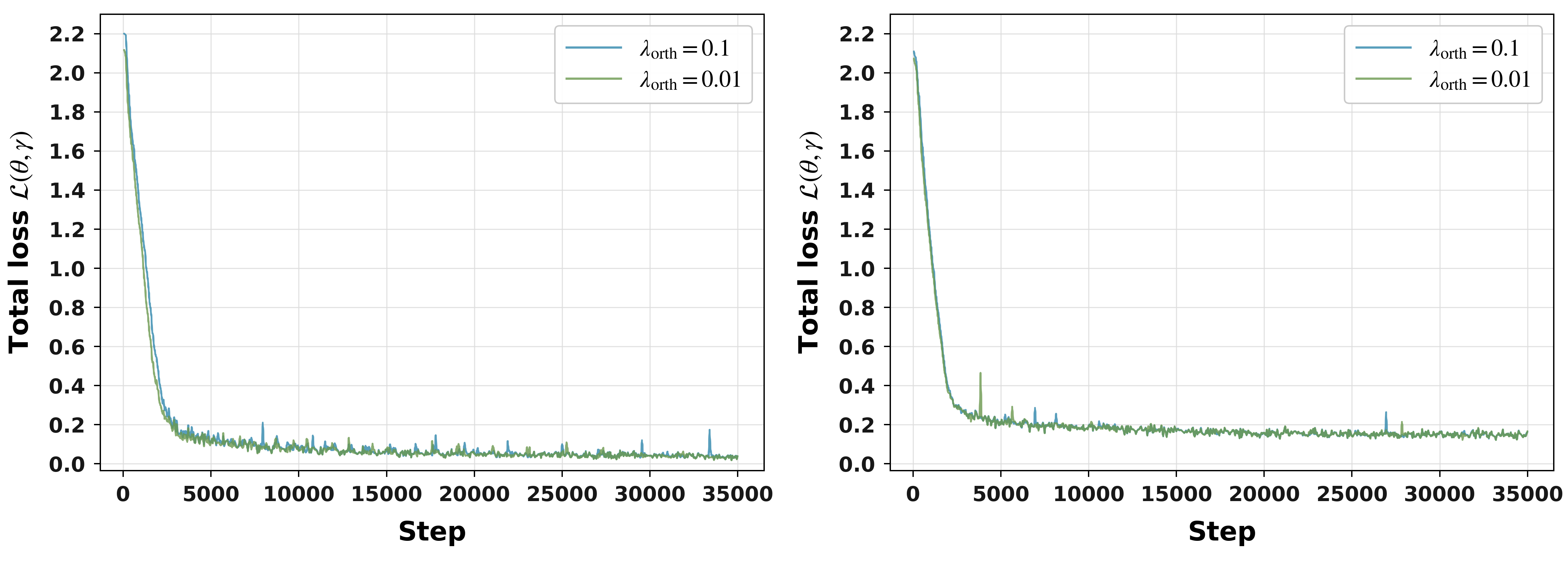}
\caption{Training loss curves. Left: Shapes3D. Right: MPI3D-toy.}
\label{Fig6}
\end{figure*}

\paragraph{Training loss curves.}
We visualize the training loss curves for different settings of the orthogonality weight $\lambda_{\mathrm{orth}}$. As shown in Figure~\ref{Fig6}, our method exhibits stable and consistent convergence across the evaluated configurations.

\begin{figure}[!t]
\centering
\includegraphics[width=\textwidth]{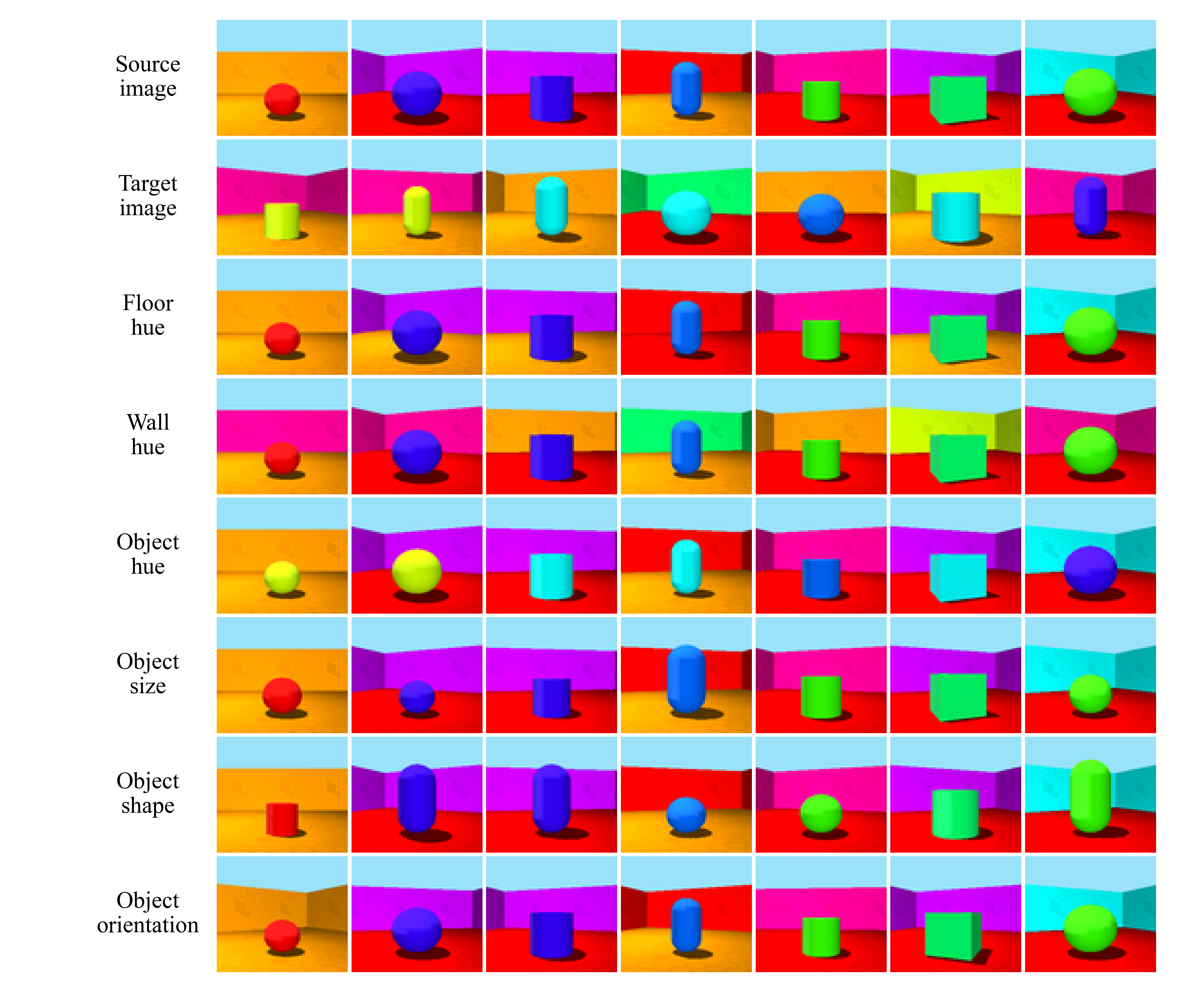}
\caption{Factor swapping on Shapes3D. Each column shows a source--target pair. The first two rows contain the source and target images. The remaining rows show generations after replacing the source token associated with the indicated factor with its target counterpart. Results cover floor hue, wall hue, object hue, object size, object shape, and orientation.}
\label{fig:shapes3d-factor-swapping}
\end{figure}

\paragraph{Additional factor-swapping results.}
Figure~\ref{fig:shapes3d-factor-swapping} extends the qualitative evaluation on Shapes3D to all six generative factors, including object shape and orientation. For each source--target pair, we replace one factor token in the source representation with its counterpart from the target representation and generate the resulting image. The examples illustrate semantic transfer through individual factor interventions.

\end{document}

%% file: introduction.tex
\section{Introduction}
\begin{wrapfigure}{r}{0.5\textwidth}
    \centering
    \vspace{-10pt}
    \includegraphics[width=\linewidth]{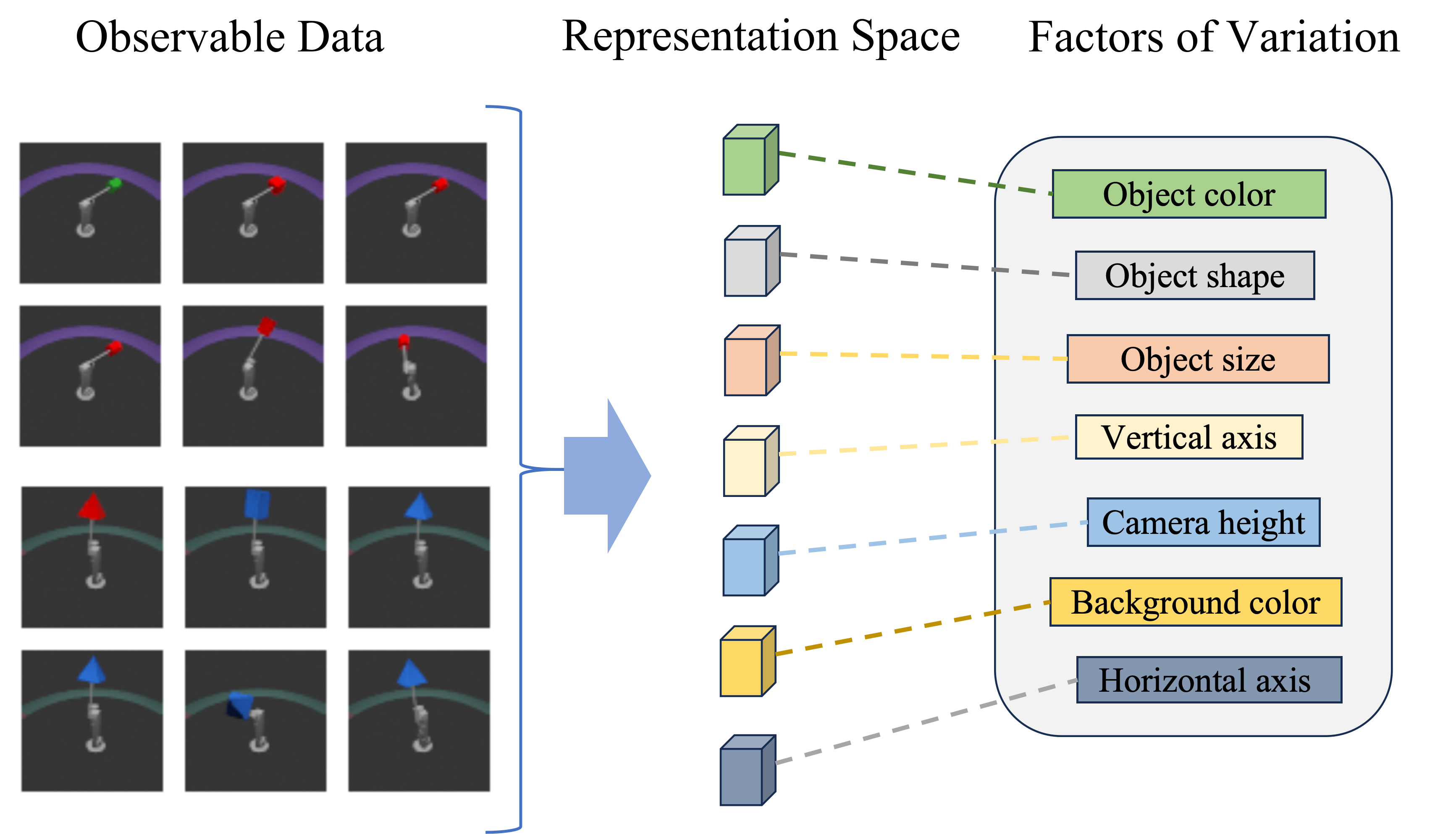}
    \vspace{-10pt}
    \caption{Illustration of the MPI3D-toy dataset \citep{gondal2019transfer}. The seven rectangles represent the underlying factors of variation in the scene, including object color, shape, size, camera height, background color, vertical axis and horizontal axis.}
    \label{Fig1}
    \vspace{-10pt}
\end{wrapfigure}

Disentangled representation learning has emerged as a fundamental objective in modern machine learning, playing a pivotal role in advances across computer vision, natural language processing (NLP), and reinforcement learning \citep{bengio2013,wang2022disentangled}. Motivated by the observation that the physical world is governed by a set of independent underlying factors, such as object shape, color, and size, this paradigm aims to encode high-dimensional observations into compact latent representations in which these semantic factors are explicitly separated. As illustrated in Figure~\ref{Fig1}, a single observation can be explained by multiple distinct factors of variation, here exemplified by seven interpretable generative factors. Such structurally disentangled representations provide a powerful inductive bias for downstream tasks, enabling controllable image synthesis and editing in vision \citep{xu2023discoscene,wang2023progressive}, facilitating style manipulation and semantic control in NLP \citep{cheng2020improving,hu2022text}, and supporting robust policy generalization and transfer in robotics \citep{higgins2018towards,wu2025discrete}. Consequently, disentangled representation learning has become an active and rapidly growing area within representation learning.

A wide range of disentangled representation learning methods have been proposed to advance this line of research. Among them, generative modeling–based methods have demonstrated particular effectiveness in learning disentangled representations from high-dimensional image data, owing to their ability to explicitly model the data generation process. On the one hand, Variational Autoencoder (VAE)–based methods, such as $\beta$-VAE \citep{higgins2016beta} and FactorVAE \citep{kim2018disentangling}, encourage disentanglement by imposing structural regularization on the latent space. However, these methods suffer from a trade-off between reconstruction fidelity and disentanglement strength \citep{chen2018isolating,burgess2018understanding}. On the other hand, Generative 
Adversarial Network (GAN)–based methods, including InfoGAN \citep{chen2016infogan} and InfoGAN-CR \citep{lin2020infogan}, pursue disentanglement from a different perspective, typically by inducing semantic structure through controllable latent variables. Despite their remarkable generative capability, GAN-based models generally lack a reversible inference mechanism between data and latent spaces, which makes them less flexible than VAEs for representation learning \citep{wang2022disentangled}.

More recently, modern generative models, particularly diffusion models, have attracted increasing attention due to their strong modeling capacity and stable training dynamics, opening new opportunities for disentangled representation learning. Existing diffusion-based methods \citep{wu2024factorized,yang2023disdiff,yang2024diffusion,jun2025disentangling} commonly introduce inductive biases to diffusion models to encourage statistical independence across latent dimensions. However, disentanglement in practice requires not only independence but also a semantic alignment criterion: \emph{each latent unit (or subspace) should correspond to a distinct, interpretable generative factor}. When semantic alignment is weak, latent dimensions may remain statistically decorrelated yet still mix factors, limiting interpretability and undermining fine-grained, factor-wise interventions.

To address these limitations, we argue that the deterministic formulation of \emph{flow matching} provides a more principled geometric basis for semantic alignment than stochastic diffusion trajectories. Flow matching learns continuous-time generative dynamics by directly matching probability flow fields, avoiding iterative denoising objectives and the associated noise-schedule design choices. It enables efficient inference through ordinary differential equation (ODE) integration. 
Building on these structural advantages, we develop a flow-matching framework that encourages semantic separation among factor representations by regularizing token-associated velocity components to reduce overlap. To the best of our knowledge, this is the first study to investigate flow matching for general disentangled representation learning. Our key contributions are three-fold:
\begin{itemize}
\item We formulate disentangled representation learning as a factor-conditioned flow in a compact latent space. This enables deterministic generation through ODE-based sampling.

\item We introduce an orthogonality regularizer on token-associated velocity components to discourage overlap and redundant factor attributions. This regularization encourages semantic separation among the learned factor representations.

\item Experiments on multiple datasets show consistent improvements in disentanglement metrics over representative baselines. Qualitative results further demonstrate improved semantic controllability and sample fidelity.
\end{itemize}

%% file: related-work.tex
\section{Related Work}
\paragraph{VAE-based methods.}
VAEs \citep{kingma2013auto} learn latent-variable generative models by maximizing the evidence lower bound (ELBO), enabling joint inference and generation. Disentanglement is typically achieved by modifying the ELBO to encourage statistical independence among latent dimensions. For example, $\beta$-VAE \citep{higgins2016beta} increases the weight of the KL term to promote a factorized prior, while FactorVAE \citep{kim2018disentangling} and $\beta$-TCVAE \citep{chen2018isolating} explicitly penalize total correlation. DIP-VAE \citep{kumar2018variational} instead aligns moments of the aggregated posterior with the prior. Overall, these methods balance reconstruction fidelity and disentanglement through objective reweighting or additional regularization.

\paragraph{GAN-based methods.}
GAN-based methods achieve disentanglement by introducing structured latent codes within adversarial training \citep{goodfellow2014generative}. InfoGAN \citep{chen2016infogan} maximizes mutual information between latent codes and generated samples to capture interpretable factors. Subsequent methods improve controllability via explicit factorization or structured latent spaces, such as separating content and style \citep{kazemi2019style,varur2025disc}. Architectural advances, e.g., StyleGAN \citep{karras2019style,karras2020analyzing}, further enable semantically meaningful control. However, these methods often suffer from limited reconstruction ability due to the difficulty of GAN inversion \citep{wang2022high}.

\paragraph{Diffusion- and flow-based methods.}
Diffusion models have emerged as a leading family of generative models, achieving state-of-the-art fidelity and mode coverage. They generate samples by reversing a gradual noising process through learned denoising or score estimation \citep{ho2020denoising,yang2023diffusion}. Unlike VAEs and GANs, diffusion-based disentanglement remains relatively underexplored, as diffusion architectures do not naturally yield compact, factorized latent representations. Consequently, existing methods typically rely on strong inductive biases to enforce factor-wise independence \citep{wu2024factorized,yang2023disdiff,yang2024diffusion,jun2025disentangling}. Specifically, \citet{wu2024factorized} decomposes images into content and mask groups to improve interpretability; \citet{yang2023disdiff} structures the denoising process to separate factors across time steps; \citet{yang2024diffusion} uses cross-attention to route distinct factors; and \citet{jun2025disentangling} enhances separation via dynamic Gaussian anchoring. 

Despite these advances, such methods remain constrained by the stochastic nature of diffusion trajectories and the complexity of iterative denoising, leading to high computational cost and limited geometric structure for precise semantic alignment. Recent work explores flow-based formulations for disentanglement. SCFlow \citep{ma2025scflow} leverages flow matching for implicit style–content disentanglement via invertible mappings. However, this method is tailored to specific settings and does not provide a general framework. In contrast, our method formulates disentanglement within a general flow-matching framework with explicit factor-level modeling.

%% file: background.tex
\section{Background}
Let $p_0(\mathbf{x})$ denote a tractable source distribution (typically a standard Gaussian) and let $p_1(\mathbf{x})$ denote the target data distribution. Flow matching is a simulation-free generative modeling framework that learns a time-dependent vector field to transport probability mass from $p_0$ to $p_1$ \citep{lipman2022flow}.

\paragraph{Probability flow ODE.}
Let $\mathbf{v}_t:\mathbb{R}^n\to\mathbb{R}^n$ be a time-dependent vector field for $t\in[0,1]$. Under mild regularity conditions, $\mathbf{v}_t$ defines a deterministic flow map $\phi_t$ as the solution to the Ordinary Differential Equation (ODE),
\begin{equation}
\label{eq:ode}
\frac{d}{dt}\phi_t(\mathbf{x})=\mathbf{v}_t(\phi_t(\mathbf{x})),\qquad \phi_0(\mathbf{x})=\mathbf{x}.
\end{equation}
Starting from an initial draw $\mathbf{x}_0\sim p_0$, the ODE generates a trajectory $\mathbf{x}_t=\phi_t(\mathbf{x}_0)$. The distribution of $\mathbf{x}_t$ is therefore the pushforward of $p_0$ through $\phi_t$,
\begin{equation}
\label{eq:pushforward}
q_t \coloneqq (\phi_t)_\# p_0.
\end{equation}
In generative modeling, we parameterize the vector field $\mathbf{v}_t$ by a neural network $\mathbf{v}_\theta(\mathbf{x},t)$ and aim to choose $\theta$ so that transporting $p_0$ to time $t=1$ matches the data distribution, i.e., $q_1 \approx p_1$. At inference time, we sample $\mathbf{x}_0\sim p_0$ and numerically integrate Eq.~\eqref{eq:ode} from $t=0$ to $1$ to obtain $\mathbf{x}_1$.

\paragraph{Conditional Flow Matching (CFM).}
Directly matching $q_1$ to $p_1$ is typically intractable. Conditional Flow Matching (CFM) \citep{lipman2022flow} instead trains $\mathbf{v}_\theta$ using supervised regression targets constructed from pairs of endpoint samples. Concretely, we draw $\mathbf{x}_0\sim p_0$ and $\mathbf{x}_1\sim p_1$, then sample a time $t\sim\mathcal{U}[0,1]$ and define an interpolation (a ``bridge'') between the endpoints,
\begin{equation}
\label{eq:cfm_bridge}
\mathbf{x}_t=\psi_t(\mathbf{x}_0,\mathbf{x}_1).
\end{equation}
The key observation is that $\psi_t$ induces a \emph{conditional} velocity along this bridge,
\begin{equation}
\label{eq:cfm_target}
\mathbf{u}_t \coloneqq \frac{d}{dt}\psi_t(\mathbf{x}_0,\mathbf{x}_1),
\end{equation}
which can be computed analytically once $\psi_t$ is chosen. CFM trains the model to predict this velocity from the intermediate point $\mathbf{x}_t$ and time $t$ by minimizing
\begin{equation}
\label{eq:cfm_loss}
\mathcal{L}_{\mathrm{CFM}}(\theta)
=
\mathbb{E}_{\mathbf{x}_0,\ \mathbf{x}_1,\ t}
\Big[
\big\|
\mathbf{v}_\theta(\mathbf{x}_t,t)-\mathbf{u}_t
\big\|_2^2
\Big].
\end{equation}
Intuitively, this objective encourages $\mathbf{v}_\theta$ to agree with the local direction of transport implied by the chosen bridge between source and data samples, without requiring explicit likelihood evaluation or solving an optimal transport problem \citep{dao2023flow}.

\paragraph{Linear interpolation.}
In this work, we use the simple linear bridge
\begin{equation}
\label{eq:linear_path}
\mathbf{x}_t = (1-t)\mathbf{x}_0 + t\mathbf{x}_1,
\end{equation}
which yields a constant velocity with respect to $t$. Differentiating Eq.~\eqref{eq:linear_path} gives
\begin{equation}
\label{eq:linear_u}
\mathbf{u}_t = \frac{d}{dt}\mathbf{x}_t = \mathbf{x}_1-\mathbf{x}_0.
\end{equation}
Plugging Eq.~\eqref{eq:linear_u} into the CFM objective in Eq.~\eqref{eq:cfm_loss} results in the training loss used in our method,
\begin{equation}
\label{eq:linear_loss}
\mathcal{L}_{\mathrm{Linear}}(\theta)
=
\mathbb{E}_{\mathbf{x}_0,\ \mathbf{x}_1,\ t}
\Big[
\big\|
\mathbf{v}_\theta(\mathbf{x}_t,t)-(\mathbf{x}_1-\mathbf{x}_0)
\big\|_2^2
\Big].
\end{equation}

%% file: method.tex
\section{Our Method}
In this work, we propose a disentangled representation learning framework based on a \emph{factor-conditioned} latent-space flow. Given an input image, we extract $N$ factor embeddings, each intended to summarize a distinct, controllable semantic attribute. We then learn a conditional flow in latent space that transports samples from a simple source prior to the data distribution. Crucially, we decompose the predicted transport velocity into $N$ token-associated components and regularize them to reduce overlap. This design encourages factor-wise separation in the learned representation and enables targeted interventions at generation time by modifying individual factor tokens used to condition the flow.

\paragraph{Overview.}
Given an image $I$, we encode it into a latent target $\mathbf z_1 = E(I)$ using a pretrained VQ-GAN encoder~\citep{esser2021taming}. This mapping induces a compact, semantically meaningful latent space and provides a stable training target, improving both efficiency and optimization stability. We then cast image generation as a \emph{transport problem} in this latent space: starting from a simple prior sample, we learn dynamics that move the latent state toward $\mathbf z_1$. To provide structured semantic cues that can steer this transport, we extract a set of $N$ factors with a trainable factor encoder $T_\gamma(\cdot)$,
\begin{equation}
S_\gamma(I)=\{\mathbf s_i\}_{i=1}^{N}=T_\gamma(I).
\end{equation}

Rather than treating the factors as a global condition, we inject them into the flow network via \emph{cross-attention}, where intermediate features attend to $S_\gamma(I)$ across multiple U-Net layers, enabling factor-aware conditioning throughout the network. This design provides a structured interface between the encoder and the flow, and has been shown to improve semantic alignment~\citep{rombach2022high,yang2024diffusion}.

We then learn a factor-conditioned vector field $\mathbf v_\theta(\mathbf z,S_\gamma(I),t)$ that governs the latent dynamics. Since the flow-matching objective constrains only the total velocity, it does not enforce factor-wise disentanglement and may lead to redundant explanations. To address this, we decompose the velocity into $N$ token-associated components and regularize them to reduce overlap, encouraging
semantic separation among factor representations. During training, we jointly optimize $\theta$ and $\gamma$ to learn disentangled and controllable representations. Figure~\ref{fig:framework} illustrates the overall framework.

\input{frame}

\subsection{Flow Matching Objective}
To train the factor-conditioned vector field $\mathbf v_\theta$, we adopt the flow matching framework and construct a regression target along a predefined interpolation path connecting a source prior sample to the data latent, conditioned on the factors $S_\gamma(I)$. Concretely, we sample $\mathbf z_0 \sim \mathcal{N}(0,\mathbf I)$ and $t\sim \mathcal{U}[0,1]$, and define the linear interpolation (as in Eq.~\eqref{eq:linear_path}):
\begin{equation}
\mathbf z_t = (1-t)\mathbf z_0 + t \mathbf z_1.
\end{equation}
This path induces a constant target velocity (cf.~Eq.~\eqref{eq:linear_u}):
\begin{equation}
\mathbf u= \frac{d \mathbf z_t}{dt} = \mathbf z_1 - \mathbf z_0.
\end{equation}
We implement $\mathbf v_\theta$ with a U-Net backbone, and inject the conditioning factors $S_\gamma(I)$ into its intermediate feature representations via cross-attention. Concretely, for a spatial feature map in the U-Net, we form queries from the spatial features and use the factors as keys and values. The cross-attention is defined as
\begin{equation}
\mathrm{Attention}(\mathbf Q,\mathbf K,\mathbf V)
=
\mathrm{softmax}\!\left(\frac{\mathbf Q\mathbf K^\top}{\sqrt{d}}\right)\mathbf V,
\end{equation}
where $d$ is the key/query dimensionality. Substituting the target velocity into the flow matching loss in Eq.~\eqref{eq:linear_loss} yields our factor-conditioned regression objective:
\begin{equation}
\label{fmt}
\mathcal{L}_{\mathrm{FM}}(\theta,\gamma)
=
\mathbb{E}_{t,\mathbf z_0,\mathbf z_1}
\Big[
\big\|\mathbf v_\theta(\mathbf z_t,S_\gamma(I),t)-(\mathbf z_1-\mathbf z_0)\big\|_2^2
\Big].
\end{equation}

\paragraph{Why factorization is needed.} 
Although the cross-attention mechanism injects encoder features into $\mathbf v_\theta$ as semantic context to associate input cues with latent transport, \emph{the flow-matching objective does not encourage disentanglement}. It constrains only the aggregate velocity to match the target direction $(\mathbf z_1-\mathbf z_0)$, without prescribing how individual factors should contribute. Consequently, different factors may redundantly explain the same transport dynamics (e.g., through channel redundancy or cross-factor leakage), resulting in ambiguous attribution and limited controllability. To obtain a factor-aligned and non-redundant decomposition, we explicitly decompose the predicted velocity into factor-specific components and introduce a regularizer that promotes diversity among them.
\subsection{Factorized Velocity via Output Attention}
We decompose the aggregate velocity field into factor-specific components and encourage them to be non-redundant via an orthogonality regularizer. In practice, we implement this factorization with an output-attention routing mask that, at each spatial location, distributes the predicted velocity across factors by assigning each factor a proportional share.

\paragraph{Factorized velocity and regularization.}
We decompose the factor-conditioned velocity field as
\begin{equation}
\mathbf v_\theta(\mathbf z_t, S_\gamma(I), t)
\;=\;
\sum_{i=1}^{N}\mathbf v_\theta^{(i)}(\mathbf z_t, S_\gamma(I), t),
\label{eq:v-aggregate}
\end{equation}
where $\mathbf v_\theta^{(i)}$ denotes the component attributed to the $i$-th factor. The flow-matching loss $\mathcal L_{\mathrm{FM}}(\theta,\gamma)$ in Eq.~\eqref{fmt} is applied to the aggregate field $\mathbf v_\theta$. The regularizer below specifies how the aggregate should be partitioned across factors.

To discourage redundant attributions, we penalize pairwise alignment between factor-specific components. For each training sample (and each sampled $t$), we flatten each component into a vector
\begin{equation}
\boldsymbol{\alpha}_i \;=\; \mathrm{vec}\!\left(\mathbf v_\theta^{(i)}(\mathbf z_t, S_\gamma(I), t)\right),
\end{equation}
and define the orthogonality loss as the average squared cosine similarity over all factor pairs,
\begin{equation}
\mathcal L_{\mathrm{orth}}
\;=\;
\frac{1}{N(N-1)}
\sum_{i\neq j}
\left(
\frac{\boldsymbol{\alpha}_i^\top \boldsymbol{\alpha}_j}{\|\boldsymbol{\alpha}_i\|_2\,\|\boldsymbol{\alpha}_j\|_2+\varepsilon}
\right)^2,
\label{eq:orth-regularizer}
\end{equation}
where $\varepsilon$ is a small constant for numerical stability. Minimizing $\mathcal L_{\mathrm{orth}}$ discourages overlap among the spatial allocations of the predicted velocity to different tokens.

\paragraph{Practical implementation via output attention.}
We instantiate $\{\mathbf v_\theta^{(i)}\}_{i=1}^{N}$ using an output-attention routing mask. Let $\mathbf h$ denote the final hidden feature map produced by the U-Net backbone that predicts $\mathbf v_\theta$. We form per-location queries and token keys as
\begin{equation}
\mathbf Q = \mathbf W_q \mathbf h, \qquad \mathbf K_i = \mathbf W_k \mathbf s_i,
\end{equation}
where $\mathbf W_q$ and $\mathbf W_k$ are learned projections. The routing weight for factor $i$ is
\begin{equation}
\mathrm{Attn}_i
=
\frac{
\exp\!\left(\frac{\mathbf Q \mathbf K_i^\top}{\sqrt{d}}\right)
}{
\sum_{j=1}^{N}
\exp\!\left(\frac{\mathbf Q \mathbf K_j^\top}{\sqrt{d}}\right)
},
\label{eq:token-attn}
\end{equation}
which satisfies $\sum_{i=1}^{N}\mathrm{Attn}_i = 1$ at each spatial location. We then define factor-specific velocities by gating the aggregate prediction:
\begin{equation}
\mathbf v_\theta^{(i)}(\mathbf z_t,S_\gamma(I),t)
=
\mathrm{Attn}_i \odot \mathbf v_\theta(\mathbf z_t,S_\gamma(I),t),
\label{eq:token-vel}
\end{equation}
where $\odot$ denotes element-wise multiplication (with $\mathrm{Attn}_i$ broadcast to match the shape of $\mathbf v_\theta$). By the simplex constraint in Eq.~\eqref{eq:token-attn}, the components sum to the aggregate field, recovering Eq.~\eqref{eq:v-aggregate}.

\paragraph{Why output attention?}
This design offers three practical advantages: (i) \emph{Factor-aligned attribution}: $\mathrm{Attn}_i$ provides an explicit, spatially varying assignment of the velocity field to each factor. (ii) \emph{Additive consistency}: the simplex constraint $\sum_i \mathrm{Attn}_i=1$ guarantees that the factor-specific components sum exactly to the aggregate prediction, preserving the flow-matching supervision. (iii) \emph{Efficiency and stability}: gating reuses the already-computed aggregate field and adds minimal overhead, while the orthogonality loss in Eq.~\eqref{eq:orth-regularizer} directly discourages redundant factor routes.

\paragraph{Overall objective.}
We optimize the combined objective
\begin{equation}
\mathcal L(\theta,\gamma)
\;=\;
\mathcal L_{\mathrm{FM}}(\theta,\gamma)
\;+\;
\lambda_{\mathrm{orth}}\,\mathcal L_{\mathrm{orth}},
\label{eq:overall-objective}
\end{equation}
where $\lambda_{\mathrm{orth}}$ controls the strength of the regularization. The full algorithm is given in Appendix C.1.

\paragraph{Choice of $\lambda_{\mathrm{orth}}$.}
In practice, we select $\lambda_{\mathrm{orth}}$ via a small validation sweep over $\{10^{-4},10^{-3},10^{-2},10^{-1},0\}$ and choose the largest value that improves factor diversity without degrading $\mathcal{L}_{\mathrm{FM}}$ or sample quality.

%% file: frame.tex
\begin{figure}[t]
\centering
\definecolor{fc1}{HTML}{E2A93B}
\definecolor{fc2}{HTML}{D45D67}
\definecolor{fc3}{HTML}{2F9585}
\definecolor{encbg}{HTML}{E6ECF7}
\definecolor{unetbg}{HTML}{FCF4DC}
\definecolor{robg}{HTML}{F8E9E9}
\definecolor{losscol}{HTML}{B23A48}
\newcommand{\cellweight}[1]{%
  \pgfmathsetmacro\ea{exp(-((\i-0.7)^2+(\j-2.5)^2)/1.3)}%
  \pgfmathsetmacro\eb{exp(-((\i-0.8)^2+(\j-0.3)^2)/1.3)}%
  \pgfmathsetmacro\ec{exp(-((\i-2.7)^2+(\j-1.6)^2)/1.3)}%
  \pgfmathsetmacro\w{ifthenelse(#1==1,\ea,ifthenelse(#1==2,\eb,\ec))/(\ea+\eb+\ec)}}
\newcommand{\velmap}[3]{%
  \draw[black!45,line width=0.4pt,fill=white] (#1,#2) rectangle ++(1,1);
  \foreach \i in {0,...,3}{\foreach \j in {0,...,3}{%
    \pgfmathsetmacro\ang{mod(\i*67+\j*41+20,360)}%
    \ifnum#3=0 \def\w{1}\def\col{black!70}\else\cellweight{#3}\def\col{fc#3}\fi
    \pgfmathsetmacro\len{0.24*\w}%
    \pgfmathsetmacro\hl{0.35+0.6*\w}%
    \ifdim\w pt<0.12pt
      \fill[black!30] (#1+0.125+0.25*\i,#2+0.125+0.25*\j) circle (0.012);
    \else
      \draw[-{Stealth[length=\hl mm]},line width=0.75pt,draw=\col]
        ($(#1+0.125+0.25*\i,#2+0.125+0.25*\j)+(\ang+180:\len/2)$) -- ++(\ang:\len);
    \fi
  }}}
\newcommand{\maskmap}[3]{%
  \foreach \i in {0,...,3}{\foreach \j in {0,...,3}{%
    \cellweight{#3}%
    \pgfmathsetmacro\op{0.03+0.87*\w}%
    \fill[fc#3,fill opacity=\op] (#1+0.25*\i,#2+0.25*\j) rectangle ++(0.25,0.25);
  }}
  \draw[black!45,line width=0.4pt] (#1,#2) rectangle ++(1,1);}
\newcommand{\ftoken}[3]{%
  \draw[#3!75!black,fill=#3!15,rounded corners=1.5pt,line width=0.5pt] (#1-0.5,#2-0.16) rectangle (#1+0.5,#2+0.16);
  \foreach \d in {-0.32,-0.15,0.32}{\fill[#3] (#1+\d,#2) circle (0.06);}
  \node[font=\scriptsize,inner sep=0pt] at (#1+0.09,#2) {$\cdots$};}
\resizebox{\linewidth}{!}{%
\begin{tikzpicture}[x=1cm,y=1cm,font=\small,
  panel/.style={rounded corners=5pt,line width=0.6pt},
  ptitle/.style={font=\small\bfseries},
  flow/.style={-{Latex[length=2mm]},line width=0.7pt,draw=black!70},
  thin/.style={-{Latex[length=1.6mm]},line width=0.55pt,draw=black!60},
  box/.style={draw=black!55,fill=white,rounded corners=2pt,line width=0.5pt,align=center,inner sep=2.5pt},
  loss/.style={draw=losscol,dashed,line width=0.7pt,rounded corners=3pt,fill=white,text=losscol!80!black,align=center,inner sep=3pt},
  note/.style={font=\footnotesize,text=black!70,align=center}]
\node at (1.0,6.45) {$I$};
\node[inner sep=0pt,draw=black!40,line width=0.4pt] (img) at (1.0,5.6)
  {\includegraphics[width=1.3cm,height=1.3cm]{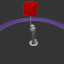}};
\draw[flow] (img.south) -- (1.0,4.38);
\draw[draw=black!55,fill=black!8,line width=0.6pt] (0.1,4.35) -- (1.9,4.35) -- (1.62,3.6) -- (0.38,3.6) -- cycle;
\node[font=\footnotesize,align=center] at (1.0,3.98) {VQ-GAN $E$\\[-2pt]{\color{black!60}(frozen)}};
\node (z1) at (1.0,3.12) {$\mathbf z_1$};
\draw[flow] (1.0,3.6) -- (z1.north);
\node[box,font=\footnotesize] (interp) at (1.0,2.15) {$\mathbf z_t=(1{-}t)\mathbf z_0+t\,\mathbf z_1$\\[1pt]{\color{losscol}$\mathbf u=\mathbf z_1-\mathbf z_0$}};
\draw[flow] (z1.south) -- (interp.north);
\node (z0) at (1.0,1.15) {$\mathbf z_0\sim\mathcal N(\mathbf 0,\mathbf I)$};
\draw[flow] (z0.north) -- (interp.south);
\node (t) at (1.0,0.45) {$t\sim\mathcal U[0,1]$};
\draw[panel,fill=encbg,draw=blue!35!black!40] (2.5,4.1) rectangle (7.6,6.65);
\node[ptitle] at (5.05,6.38) {Factor encoder $T_\gamma$};
\draw[flow] (img.east |- 0,5.3) -- (2.75,5.3);
\draw[draw=blue!50!black,fill=blue!22,line width=0.6pt] (2.75,6.0) -- (3.65,5.7) -- (3.65,4.9) -- (2.75,4.6) -- cycle;
\node at (3.2,5.3) {CNN};
\draw[flow] (3.65,5.3) -- (3.92,5.3);
\draw[draw=blue!50!black,fill=blue!12,line width=0.6pt,rounded corners=2pt] (3.92,4.75) rectangle (4.42,5.85);
\node at (4.17,5.3) {FC};
\foreach \y/\c/\l in {5.9/fc1/1,5.45/fc2/2,4.7/fc3/N}
  {\draw[thin] (4.42,5.3) -- (5.1,\y);
   \ftoken{5.6}{\y}{\c}
   \node at (6.4,\y) {$\mathbf s_{\l}$};}
\node at (5.6,5.12) {$\vdots$};
\node[note] at (5.6,4.3) {$S_\gamma(I)$};
\draw[decorate,decoration={brace,amplitude=4pt},line width=0.6pt,black!60] (6.75,6.08) -- (6.75,4.52);
\draw[line width=0.7pt,black!70] (6.92,5.3) -- (7.3,5.3) -- (7.3,3.75);
\draw[panel,fill=unetbg,draw=orange!50!black!40] (2.5,-0.05) rectangle (7.6,3.5);
\foreach \x/\hh in {3.2/2.1,4.0/1.5,4.95/0.9,5.9/1.5,6.7/2.1}
  {\draw[draw=orange!65!black,fill=orange!22,rounded corners=1.5pt,line width=0.6pt]
     (\x-0.16,2.0-\hh/2) rectangle (\x+0.16,2.0+\hh/2);}
\foreach \a/\b in {3.2/4.0,4.0/4.95,4.95/5.9,5.9/6.7}
  {\draw[thin] (\a+0.16,2.0) -- (\b-0.16,2.0);}
\draw[black!55,dashed,line width=0.7pt,-{Latex[length=1.4mm]}] (3.2,0.95) -- (3.2,0.8) -- (6.7,0.8) -- (6.7,0.95);
\draw[black!55,dashed,line width=0.7pt,-{Latex[length=1.4mm]}] (4.0,1.25) -- (4.0,1.1) -- (5.9,1.1) -- (5.9,1.25);
\node[ptitle] at (5.05,0.2) {Flow-matching U-Net};
\node[note] at (5.15,0.53) {CA: $\mathbf K,\mathbf V\!\leftarrow\!S_\gamma(I)$, $\mathbf Q\!\leftarrow$ features};
\draw[line width=0.7pt,black!70] (4.0,3.75) -- (7.3,3.75);
\foreach \x/\hh in {4.0/1.5,4.95/0.9,5.9/1.5}
  {\draw[flow] (\x,3.75) -- (\x,3.33);
   \node[draw=orange!70!black,fill=orange!40,rounded corners=2pt,minimum width=0.6cm,minimum height=0.4cm,inner sep=1pt,font=\footnotesize] at (\x,3.1) {CA};
   \draw[orange!65!black,line width=0.6pt] (\x,2.9) -- (\x,2.0+\hh/2);}
\draw[flow] (interp.east |- 0,2.32) -- (3.04,2.32);
\draw[flow] (t.east) -| (2.75,1.2) -- (3.04,1.2);
\node[draw=orange!70!black,fill=orange!35,rounded corners=1pt,minimum width=0.3cm,minimum height=1.25cm,inner sep=0pt] (h) at (7.28,1.98) {};
\node at (7.28,2.85) {$\mathbf h$};
\draw[thin] (6.86,1.98) -- (h.west);
\draw[panel,fill=robg,draw=red!40!black!40] (7.95,-0.05) rectangle (15.45,6.65);
\node[ptitle] at (11.7,6.38) {Factorized velocity via output attention};
\node[box] (key) at (8.4,4.6) {$\mathbf K_i$};
\node[box] (qry) at (8.4,2.35) {$\mathbf Q$};
\draw[flow] (7.3,4.6) -- (key.west);
\draw[flow] (h.east |- 0,2.35) -- (qry.west);
\node[font=\scriptsize,text=black!60] at (8.4,5.02) {$\mathbf W_k\mathbf s_i$};
\node[font=\scriptsize,text=black!60] at (8.4,1.93) {$\mathbf W_q\mathbf h$};
\node[box,font=\footnotesize] (soft) at (9.7,3.5) {$\mathrm{softmax}_i$\\[2pt]$\bigl(\mathbf Q\mathbf K_i^{\top}\!/\sqrt d\,\bigr)$\\[2pt]{\color{black!65}$\textstyle\sum_i\mathrm{Attn}_i=\mathbf 1$}};
\draw[thin] (key.south) -- (soft.north west);
\draw[thin] (qry.north) -- (soft.south west);
\velmap{11.65}{0.3}{0}
\draw[flow] (h.east |- 0,1.6) -- (8.2,1.6) |- (11.65,0.8);
\node[box,font=\footnotesize] at (9.4,0.8) {output conv};
\node[anchor=south east] at (11.63,0.85) {$\mathbf v_\theta$};
\node[font=\footnotesize] at (11.3,6.0) {$\mathrm{Attn}_i$};
\node[font=\footnotesize] at (13.0,6.0) {$\mathbf v_\theta^{(i)}$};
\draw[line width=0.7pt,black!70] (12.15,1.3) -- (12.15,5.3);
\foreach \k/\y in {1/5.3,2/3.95,3/2.35}
  {\draw[thin] (soft.east) -- (10.8,\y);
   \maskmap{10.8}{\y-0.5}{\k}
   \draw[thin] (11.8,\y) -- (12.02,\y);
   \node[circle,draw=black!70,fill=white,inner sep=0pt,minimum size=0.26cm,font=\scriptsize] at (12.15,\y) {$\odot$};
   \draw[thin] (12.28,\y) -- (12.5,\y);
   \velmap{12.5}{\y-0.5}{\k}}
\node at (11.3,3.2) {$\vdots$};
\node at (13.0,3.2) {$\vdots$};
\draw[decorate,decoration={brace,amplitude=4pt},line width=0.6pt,losscol] (13.7,5.8) -- (13.7,1.85);
\node[loss] at (14.55,3.82) {$\mathcal L_{\mathrm{orth}}$\\[2pt]{\footnotesize pairwise}\\[-1pt]{\footnotesize $\cos^2$}};
\node[loss] (lfm) at (14.55,0.8) {$\mathcal L_{\mathrm{FM}}$\\[1pt]{\footnotesize $\|\mathbf v_\theta-\mathbf u\|_2^2$}};
\draw[flow,draw=losscol] (12.65,0.8) -- (lfm.west);
\draw[flow,draw=losscol,dashed] (interp.west |- 0,1.98) -- (-0.45,1.98) |- (3.0,-0.3) -| (lfm.south);
\node[font=\small,text=losscol,anchor=east] at (-0.47,0.9) {$\mathbf u$};
\node[anchor=north] at (9.6,-0.3) {Objective: $\mathcal L=\mathcal L_{\mathrm{FM}}+\lambda_{\mathrm{orth}}\,\mathcal L_{\mathrm{orth}}$};
\end{tikzpicture}%
}
\caption{Training pipeline of the proposed framework. A frozen VQ-GAN encoder $E$ maps $I$ to $\mathbf z_1$, which is linearly interpolated with Gaussian noise $\mathbf z_0$ to obtain $\mathbf z_t$. Factor tokens $S_\gamma(I)=\{\mathbf s_i\}_{i=1}^N$ from $T_\gamma$ condition the U-Net through cross-attention (CA). Output attention computes soft routing masks $\mathrm{Attn}_i$ from $\mathbf h$ and the tokens (Eqs.~(17)--(18)). These masks decompose the predicted velocity into factor-specific components $\mathbf v_\theta^{(i)}$ that sum to $\mathbf v_\theta$ (Eqs.~(14) and (19)). $\mathcal L_{\mathrm{FM}}$ supervises the aggregate velocity, while $\mathcal L_{\mathrm{orth}}$ discourages overlap between components. Masks and velocity fields are schematic; dots indicate small velocity magnitudes.}
\label{fig:framework}
\end{figure}

%% file: experiments.tex
\section{Experiments}
We present qualitative and quantitative results that demonstrate our method's effectiveness on synthetic and real-world data. All experiments are conducted on a Linux server equipped with $8\times$ NVIDIA RTX 4090 GPUs. The code will be released at \url{https://github.com/jinjinchi/LDR-FM}.

\begin{table*}[t]
\centering
\caption{Quantitative comparison on Cars3D, Shapes3D, and MPI3D-toy datasets using FactorVAE score and DCI-D (mean$\pm$std).}
\label{tab:quant_comparison}
\small
{\fontsize{6.5pt}{8pt}\selectfont
\setlength{\tabcolsep}{5.5pt}
\begin{tabular}{@{} c l cc cc cc @{}}
\toprule
\multirow{2}{*}{{}} & \multirow{2}{*}{{Method}} &
\multicolumn{2}{c}{{Cars3D}} &
\multicolumn{2}{c}{{Shapes3D}} &
\multicolumn{2}{c}{{MPI3D-toy}} \\
\cmidrule(lr){3-4}\cmidrule(lr){5-6}\cmidrule(lr){7-8}
& & {FactorVAE }$\uparrow$ & {DCI-D }$\uparrow$ &
{FactorVAE }$\uparrow$ & {DCI-D }$\uparrow$ &
{FactorVAE }$\uparrow$ & {DCI-D }$\uparrow$ \\
\midrule

\multirow{3}{*}{\rotatebox{90}{{VAE}}}
& FactorVAE     & $0.906\pm0.052$ & $0.161\pm0.019$ & $0.840\pm0.066$ & $0.611\pm0.101$ & $0.152\pm0.025$ & $0.240\pm0.051$ \\
& $\beta$-TCVAE    & $0.855\pm0.082$ & $0.140\pm0.019$ & $0.873\pm0.074$ & $0.613\pm0.114$ & $0.179\pm0.017$ & $0.237\pm0.056$ \\
& DAVA             & $0.940\pm0.010$ & $0.230\pm0.040$ & $0.820\pm0.030$ & $0.780\pm0.030$ & $0.410\pm0.040$ & $0.300\pm0.030$ \\
\midrule

\multirow{3}{*}{\rotatebox{90}{{GAN}}}
& ClosedForm           & $0.873\pm0.036$ & $0.243\pm0.048$ & $0.951\pm0.021$ & $0.525\pm0.078$ & $0.523\pm0.056$ & $0.318\pm0.014$ \\
& GANSpace       & $0.932\pm0.018$ & $0.209\pm0.031$ & $0.788\pm0.091$ & $0.284\pm0.034$ & $0.465\pm0.036$ & $0.229\pm0.042$ \\
& DisCo-GAN              & $0.855\pm0.074$ & $0.271\pm0.037$ & $0.877\pm0.031$ & $0.708\pm0.048$ & $0.371\pm0.030$ & $0.292\pm0.024$ \\
\midrule

\multirow{4}{*}{\rotatebox{90}{{Diffusion}}}
& DisDiff         & $0.976\pm0.018$ & $0.232\pm0.019$ & $0.902\pm0.043$ & $0.723\pm0.013$ & $0.617\pm0.070$ & $0.337\pm0.057$ \\
& FDAE                & $0.918\pm0.027$ & $0.232\pm0.041$ & $0.987\pm0.023$ & $0.917\pm0.038$ & $0.779\pm0.036$ & $0.528\pm0.042$ \\
& EncDiff & $0.773\pm0.060$ & $0.279\pm0.022$ & $0.999\pm0.000$ & $0.969\pm0.030$ & $0.872\pm0.049$ & $0.685\pm0.044$ \\
& DyGA          & $0.941\pm0.002$ & $0.414\pm0.013$ & $\mathbf{1.000\pm0.000}$ & $0.938\pm0.001$ & $0.930\pm0.004$ & $0.627\pm0.002$ \\
\midrule

& \textbf{Ours}  & $\mathbf{0.980\pm0.011}$ & $\mathbf{0.487\pm0.034}$ & $\mathbf{1.000\pm0.000}$ & $\mathbf{0.977\pm0.032}$ & $\mathbf{0.931\pm0.017}$ & $\mathbf{0.710\pm0.022}$ \\
\bottomrule
\end{tabular}
}
\end{table*}

\subsection{Experimental Setup}

\paragraph{Datasets.}
We evaluate on standard disentanglement datasets: Cars3D~\citep{reed2015deep}, Shapes3D~\citep{3dshapes18}, and MPI3D-toy~\citep{gondal2019transfer}. Cars3D comprises 3D-rendered car images with viewpoint-related factors. Shapes3D consists of procedurally rendered 3D objects with multiple controlled generative factors (e.g., shape, size, orientation, and color). MPI3D-toy contains images rendered in a controlled environment with annotated factors of variation. To assess performance on real-world imagery, we additionally report results on CelebA \citep{liu2015deep}, a large-scale face dataset with attribute annotations.

\paragraph{Baselines.}
We compare against three families of baselines:
\emph{VAE-based}: FactorVAE~\citep{kim2018disentangling}, $\beta$-TCVAE~\citep{chen2018isolating}, and DAVA~\citep{estermanndava};
\emph{GAN-based}: ClosedForm~\citep{shen2021closed}, GANSpace~\citep{harkonen2020ganspace}, and DisCo-GAN~\citep{renlearning};
\emph{Diffusion-based}: DisDiff~\citep{yang2023disdiff}, FDAE~\citep{wu2024factorized}, EncDiff~\citep{yang2024diffusion}, and DyGA~\citep{jun2025disentangling}.
We use official implementations when available, noting that several baselines also employ pre-trained encoders in their original setups.

\paragraph{Evaluation metrics.}
On synthetic datasets, we report the FactorVAE score~\citep{kim2018disentangling} and the disentanglement component of DCI~\citep{eastwood2018framework}, denoted DCI-D. These metrics assess factor-to-coordinate alignment and the specificity of individual representation coordinates, respectively. On CelebA, we report Total AUROC Difference (TAD)~\citep{yeats2022nashae}, which measures attribute alignment through the gap between the two highest per-coordinate AUROC scores for each evaluated attribute. We use Fr\'echet Inception Distance (FID)~\citep{heusel2017gans} to evaluate generation quality on CelebA. Following common practice, all metrics except FID are computed on PCA-reduced latent representations when the latent space is high-dimensional.

\paragraph{Implementation details.}
We use a batch size of 128 across all datasets. Unless otherwise stated, our model uses $N=20$ latent factor tokens and an orthogonality weight of $\lambda_{\mathrm{orth}}=0.01$. During sampling, we solve the associated ODE using the adaptive-step Dopri5 solver~\citep{Akhtar2025TuningNeuralODE}. Architecture details and additional experimental results are provided in Appendices C.2 and C.3, respectively.

\subsection{Comparison with State-of-the-Art Methods}

\paragraph{Results: Cars3D, Shapes3D, and MPI3D-toy.}
Table 1 reports disentanglement results on three synthetic datasets using the FactorVAE score and DCI-D. Our method achieves the highest mean DCI-D and the highest or tied highest mean FactorVAE score on all three datasets. The improvements over VAE- and GAN-based baselines are substantial. On MPI3D-toy, our method achieves a FactorVAE score of 0.931 and a DCI-D score of 0.710, exceeding the strongest VAE baseline, DAVA (0.410/0.300), and the strongest GAN baseline, ClosedForm (0.523/0.318). Compared with diffusion-based baselines, the gains in DCI-D are smaller but consistent, while FactorVAE scores are matched or improved. Since diffusion models already provide strong generative backbones, further improvements in disentanglement can benefit from explicitly encouraging semantic structure. Our method promotes such structure through factor-aligned representations, improving DCI-D without degrading the FactorVAE score.

\paragraph{Results: CelebA.}
Table~\ref{lab:lab2} reports disentanglement-related behavior (TAD$\uparrow$) and sample quality (FID$\downarrow$) on the CelebA dataset. Our method achieves the highest mean TAD and the lowest FID among the evaluated methods, indicating stronger attribute alignment and improved generation quality. These results demonstrate that the benefits of our method extend beyond synthetic datasets to real-world face images.

\begin{table}[h]
\centering
\small
{\fontsize{7.5pt}{9pt}\selectfont
\setlength{\tabcolsep}{5.5pt}
\caption{Attribute alignment (TAD) and generation quality (FID) on CelebA (mean$\pm$std).}
\label{lab:lab2}
\begin{tabular}{lcccccc}
\toprule
\textbf{Metric} & FactorVAE & DisDiff & FDAE & EncDiff & DyGA & \textbf{Ours} \\
\midrule
TAD $\uparrow$ & $0.137 \pm 0.043$ & $0.305 \pm 0.010$ & $0.275 \pm 0.104$ & $0.638 \pm 0.008$ & $0.905 \pm 0.021$ & $\mathbf{0.913 \pm 0.012}$ \\
FID $\downarrow$ & $96.3 \pm 1.6$ & $18.2 \pm 2.1$ & $18.8 \pm 1.5$ & $14.8 \pm 2.3$ & $12.1 \pm 0.9$ & $\mathbf{7.8 \pm 0.1}$ \\
\bottomrule
\end{tabular}
}
\end{table}

\subsection{Visualization Results}
\begin{figure}[t]
\centering
\includegraphics[width=\textwidth]{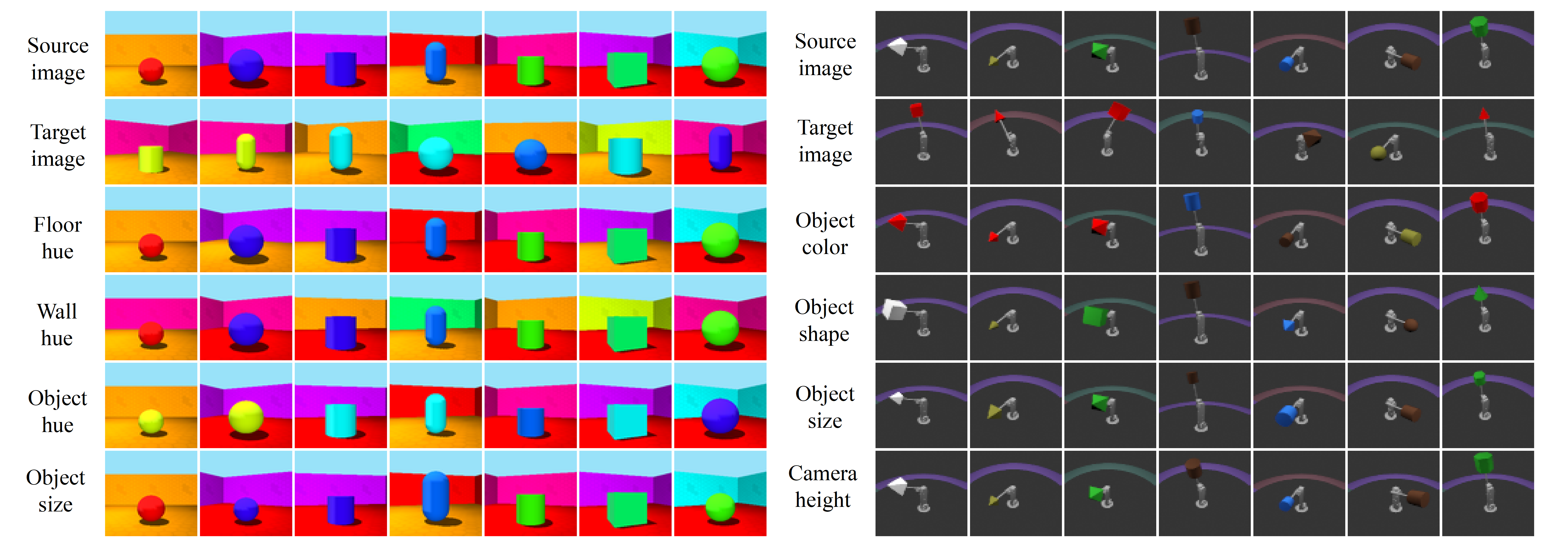}
\caption{Factor swapping on Shapes3D (left) and MPI3D-toy (right). The top two rows show source and target images. Subsequent rows show images generated after replacing one source factor token with the corresponding target token; each row corresponds to a different exchanged factor.}
\label{Fig3}
\end{figure}

A simple yet effective way to evaluate whether the feature extractor learns \emph{well-disentangled} representations is to actively intervene on the inferred factors and inspect the resulting generations. If changing a single factor induces a predictable, semantically coherent modification in the output while leaving other attributes largely unchanged, this indicates that the representation separates underlying generative factors \citep{bengio2013}. We therefore perform a \emph{factor swapping} experiment on Shapes3D and MPI3D-toy. Given a source image and a target image, we encode each into its set of factors, replace one source factor with the corresponding target factor, and synthesize an image conditioned on the modified representation. Figure~\ref{Fig3} shows that the corresponding attributes, such as wall hue and object shape, are transferred from the target image while the source image's other attributes are preserved. These results provide qualitative evidence of disentanglement and factor-level controllability. Additional factor-swapping results for object shape and orientation on Shapes3D are provided in Figure \ref{fig:shapes3d-factor-swapping} in the Appendix.

\subsection{Ablation Study}
We conduct ablation studies to assess the contributions of the conditioning mechanism and the orthogonality regularizer to disentanglement. Cross-attention allows spatial features in the U-Net to attend to individual factor tokens, providing a structured interface between the factor encoder and the flow network. To examine whether this interaction benefits disentanglement, we replace cross-attention conditioning with adaptive group normalization (\emph{AdaGN conditioning}). In addition, the flow-matching objective supervises the aggregate velocity without explicitly constraining how individual factors contribute to it. This motivates the orthogonality regularizer, which discourages redundancy among factor-specific velocity components. We therefore evaluate a variant without this regularizer (\emph{w/o $\mathcal{L}_{\mathrm{orth}}$}), setting $\lambda_{\mathrm{orth}}=0$, to assess whether explicit factor separation provides additional benefits beyond the base flow-matching objective. Table~\ref{tab:ablation} compares both variants with the full model on Cars3D, Shapes3D, and MPI3D-toy, using $N=20$.
\begin{table}[t]
\centering
\small
{\fontsize{7pt}{8.5pt}\selectfont
\setlength{\tabcolsep}{4pt}
\caption{Ablation of the conditioning mechanism and orthogonality regularizer (mean$\pm$std).}
\label{tab:ablation}
\begin{tabular}{lcccccc}
\toprule
\multirow{2}{*}{\textbf{Variant}} 
& \multicolumn{2}{c}{Cars3D} 
& \multicolumn{2}{c}{Shapes3D} 
& \multicolumn{2}{c}{MPI3D-toy} \\
\cmidrule(lr){2-3} \cmidrule(lr){4-5} \cmidrule(lr){6-7}
& FactorVAE $\uparrow$ & DCI-D $\uparrow$
& FactorVAE $\uparrow$ & DCI-D $\uparrow$
& FactorVAE $\uparrow$ & DCI-D $\uparrow$ \\
\midrule
AdaGN
& $0.954 \pm 0.028$ & $0.308 \pm 0.084$
& $0.865 \pm 0.024$ & $0.684 \pm 0.040$
& $0.695 \pm 0.039$ & $0.440 \pm 0.040$ \\
w/o $\mathcal{L}_{\mathrm{orth}}$
& $0.914 \pm 0.002$ & $0.448 \pm 0.046$
& $0.999 \pm 0.001$ & $0.946 \pm 0.021$
& $0.804 \pm 0.073$ & $0.637 \pm 0.041$ \\

\textbf{Full model}
& $\mathbf{0.980 \pm 0.011}$ & $\mathbf{0.487 \pm 0.034}$
& $\mathbf{1.000 \pm 0.000}$ & $\mathbf{0.977 \pm 0.032}$
& $\mathbf{0.931 \pm 0.017}$ & $\mathbf{0.710 \pm 0.022}$ \\
\bottomrule
\end{tabular}
}
\end{table}

\begin{table}[t]
\centering
\small
{\fontsize{7.5pt}{9pt}\selectfont
\setlength{\tabcolsep}{5.5pt}
\caption{Sensitivity analysis with respect to the number of latent factor tokens $N$ on the Cars3D, Shapes3D, and MPI3D-toy datasets (mean$\pm$std).}
\label{tab:sensitivity}

\begin{tabular}{ccccccc}
\toprule
 & \multicolumn{2}{c}{Cars3D} & \multicolumn{2}{c}{Shapes3D} & \multicolumn{2}{c}{MPI3D-toy} \\
\cmidrule(lr){2-3} \cmidrule(lr){4-5} \cmidrule(lr){6-7}
$N$ & FactorVAE $\uparrow$ & DCI-D $\uparrow$ 
    & FactorVAE $\uparrow$ & DCI-D $\uparrow$ 
    & FactorVAE $\uparrow$ & DCI-D $\uparrow$ \\
\midrule
5  & $0.909 \pm 0.059$ & $0.195 \pm 0.086$ 
   & $0.795 \pm 0.045$ & $0.710 \pm 0.074$ 
   & $0.443 \pm 0.040$ & $0.282 \pm 0.069$ \\
10 & $0.964 \pm 0.013$ & $0.431 \pm 0.034$ 
   & $1.000 \pm 0.000$ & $0.973 \pm 0.020$ 
   & $0.907 \pm 0.017$ & $0.649 \pm 0.024$ \\
15 & $0.934 \pm 0.027$ & $0.521 \pm 0.045$ 
   & $1.000 \pm 0.000$ & $0.962 \pm 0.032$ 
   & $0.926 \pm 0.013$ & $0.687 \pm 0.004$ \\
20 & $0.980 \pm 0.011$ & $0.487 \pm 0.034$ 
   & $1.000 \pm 0.000$ & $0.977 \pm 0.032$ 
   & $0.931 \pm 0.017$ & $0.710 \pm 0.022$ \\
30 & $0.963 \pm 0.036$ & $0.582 \pm 0.040$ 
   & $1.000 \pm 0.000$ & $0.946 \pm 0.037$ 
   & $0.943 \pm 0.035$ & $0.757 \pm 0.045$ \\
\bottomrule
\end{tabular}
}
\end{table}

\paragraph{Conditioning mechanism.}
Replacing cross-attention with AdaGN reduces both metrics on all three datasets. The DCI-D score decreases from $0.487$ to $0.308$ on Cars3D, from $0.977$ to $0.684$ on Shapes3D, and from $0.710$ to $0.440$ on MPI3D-toy. These results support cross-attention as an effective interface between factor tokens and the latent flow network.

\paragraph{Orthogonality regularization.}
Adding $\mathcal{L}_{\mathrm{orth}}$ improves both metrics across the three datasets. On MPI3D-toy, the FactorVAE score increases from $0.804$ to $0.931$, and DCI-D increases from $0.637$ to $0.710$. This ablation supports the contribution of the regularizer to learning more disentangled representations beyond cross-attention conditioning alone.

\subsection{Sensitivity to the Number of Tokens $N$}
We vary $N\in\{5,10,15,20,30\}$ to examine how token capacity affects disentanglement and report the results in Table~\ref{tab:sensitivity}. On Shapes3D and MPI3D-toy, increasing $N$ from $5$ to $10$ makes a dedicated token for each ground-truth factor possible and yields substantial gains. Cars3D also benefits despite $N=5$ already exceeding its three factors, suggesting that capacity requirements are not determined solely by factor count. Further increases in $N$ improve both the FactorVAE score and DCI-D on MPI3D-toy, whereas DCI-D varies non-monotonically on Shapes3D and Cars3D.

These scores alone do not reveal how extra tokens are used under over-specification. Output attention in Eqs.~\eqref{eq:token-attn} and \eqref{eq:token-vel} routes velocity softly without explicit pruning. Extra tokens may receive little output attention or participate in encoding existing factors. To assess the effects of over-specification on the learned representations, we report DCI informativeness (DCI-I) and completeness (DCI-C) for $N\in\{10,20,30\}$ in Appendix Table~\ref{tab:dci_ic_sensitivity}. DCI-I measures factor predictability, while DCI-C measures the concentration of each factor's predictive importance across representation components. 

DCI-I is maintained or improved across the three datasets, whereas DCI-C decreases on Shapes3D and MPI3D-toy and increases on Cars3D. On MPI3D-toy, increasing $N$ from $10$ to $30$ raises DCI-D from $0.649$ to $0.757$, while DCI-C falls from $0.609$ to $0.573$. This pattern is consistent with the definition of DCI-D. The score down-weights unimportant components and can remain high when multiple components encode the same factor, provided each remains factor-specific. Additional token capacity can thus preserve or improve disentanglement and predictability while making factor information less concentrated.

\subsection{Downstream Tasks}
To assess the practical utility of our learned representations under limited supervision, we follow the evaluation protocol of \citet{jun2025disentangling} and evaluate downstream classification on Shapes3D and MPI3D-toy using gradient-boosted trees (GBT). We train classifiers on the learned representations with different numbers of labeled samples. Let $\mathrm{Acc}_{1000}$ and $\mathrm{Acc}_{100}$ denote the accuracies obtained using 1,000 and 100 training samples, respectively, and let $\mathrm{Acc}$ denote the reference accuracy obtained using 10,000 training samples. We report $\mathrm{Acc}_{1000}/\mathrm{Acc}$ and $\mathrm{Acc}_{100}/\mathrm{Acc}$, which measure the fraction of each method's reference accuracy retained under limited supervision. This evaluation assesses relative sample efficiency when labeled data are scarce.

We compare our method with DisDiff, FDAE, EncDiff, and DyGA. Table~\ref{tab:efficiency} shows that our method achieves the highest mean accuracy ratios at both sample budgets on both datasets. For example, with 100 labeled samples on Shapes3D, our method retains $89.0\%$ of its reference accuracy, compared with $86.3\%$ for DyGA. These results indicate better retention of downstream classification performance as the number of labeled samples decreases, supporting the relative sample efficiency of our learned representations under this evaluation protocol.

\begin{table}[t]
\small
 \centering
{\fontsize{8.5pt}{10pt}\selectfont
\setlength{\tabcolsep}{5.5pt}
   
    \caption{Relative downstream accuracy with limited training labels on Shapes3D and MPI3D-toy. $\mathrm{Acc}$ denotes accuracy with 10,000 training examples. Higher ratios indicate better retention of this accuracy.}
    \label{tab:efficiency}
    \begin{tabular}{lcccc}
        \toprule
        \multirow{2}{*}{\textbf{Method}} & \multicolumn{2}{c}{Shapes3D} & \multicolumn{2}{c}{MPI3D-toy} \\
        \cmidrule(lr){2-3} \cmidrule(lr){4-5}
        & $\text{Acc}_{1000} / \text{Acc}$ & $\text{Acc}_{100} / \text{Acc}$ & $\text{Acc}_{1000} / \text{Acc}$ & $\text{Acc}_{100} / \text{Acc}$ \\
        \midrule
        DisDiff & $0.928 \pm 0.001$ & $0.732 \pm 0.002$ & $0.862 \pm 0.001$ & $0.700 \pm 0.002$ \\
        FDAE    & $0.983 \pm 0.004$ & $0.746 \pm 0.004$ & $0.847 \pm 0.011$ & $0.745 \pm 0.012$ \\
        EncDiff & $0.975 \pm 0.000$ & $0.772 \pm 0.002$ & $0.853 \pm 0.005$ & $0.701 \pm 0.001$ \\
         DyGA & $0.990 \pm 0.000$ & $0.863 \pm 0.001$ & $0.888 \pm 0.002$ & $0.757 \pm 0.001$ \\
        \midrule
        \textbf{Ours} & $\mathbf{0.994 \pm 0.001}$ & $\mathbf{0.890 \pm 0.003}$ & $\mathbf{0.899 \pm 0.002}$ & $\mathbf{0.759 \pm 0.012}$ \\
        \bottomrule
    \end{tabular}
    }
\end{table}